\pdfoutput=1
\documentclass[11pt]{article}

\usepackage[]{CoNLL2021}

\usepackage{times}
\usepackage{latexsym}
\usepackage{graphicx}
\usepackage{amsmath}
\usepackage{multirow}

\usepackage[T1]{fontenc}
\usepackage[utf8]{inputenc}

\usepackage{microtype}

\title{Imposing Relation Structure in Language-Model Embeddings \\ Using Contrastive Learning}
\author{Christos Theodoropoulos \\
  KU Leuven \\
  \texttt{\small christos.theodoropoulos@kuleuven.be} \\\And
  James Henderson \\
  Idiap Research Institute \\
  \texttt{\small james.henderson@idiap.ch} \\\AND 
  Andrei C. Coman \\
  EPFL, Idiap Research Institute \\
  \texttt{\small andrei.coman@idiap.ch} \\\And
  Marie-Francine Moens \\
  KU Leuven \\
  \texttt{\small sien.moens@kuleuven.be}}

\begin{document}
\maketitle
\begin{abstract}
Though language model text embeddings have revolutionized NLP research, their ability to capture high-level semantic information, such as relations between entities in text, is limited.  In this paper, we propose a novel contrastive learning framework that trains sentence embeddings to encode the relations in a graph structure.  Given a sentence (unstructured text) and its graph, we use contrastive learning to impose relation-related structure on the token-level representations of the sentence obtained with a CharacterBERT \cite{el2020characterbert} model.  The resulting relation-aware sentence embeddings achieve state-of-the-art results on the relation extraction task using only a simple KNN classifier, thereby demonstrating the success of the proposed method.  Additional visualization by a tSNE analysis shows the effectiveness of the learned representation space compared to baselines.  Furthermore, we show that we can learn a different space for named entity recognition, again using a contrastive learning objective, and demonstrate how to successfully combine both representation spaces in an entity-relation task.
\vspace{-1mm}
\end{abstract}

\section{Introduction}
Pretrained language models (LMs), such as BERT \cite{devlin2018bert}, RoBERTa \cite{liu2019roberta} and GPT-3 \cite{NEURIPS2020_1457c0d6}, capture contextualized information effectively and are used in a wide variety of natural language processing (NLP) tasks.  They have revolutionized NLP research. The main mechanism of these models is multi-head self-attention \cite{NIPS2017_3f5ee243}, which enables capturing patterns of semantic and syntactic interest in text. However, their ability to encapsulate high level semantic information, such as relations in the text, and domain-specific knowledge, is limited because they are trained on very large corpora using the main objectives of language modeling. In many NLP tasks, pretrained LM embeddings are used as model input. A common strategy is to concatenate the embeddings that are extracted from different LMs and let the model decide which part of the information is useful for the task. This empirical approach does not provide strong intuition and results in poor explainability capabilities because most of the task-specific models are black boxes.

\par
In this study, we present a novel contrastive learning (CL) framework to leverage the embedding space of CharacterBERT and impose a relation structure on the embeddings. The proposed framework receives a sentence and a graph that represents the text relations in a structured way, and the CL paradigm is applied to impose this structure on the token embeddings of the CharacterBERT text encoder. Different graph formulations that represent the text relations are explored. The main goal is to create a common embedding space where relations can be easily detected. To evaluate progress towards this goal, we use the ADE dataset  \cite{gurulingappa2012development}, which is widely used for relation extraction (RE) \cite{zhao2005extracting}; \cite{jiang2007systematic}; \cite{sun2011semi}; \cite{plank2013embedding} and named entity recognition (NER) tasks \cite{curran2003language}; \cite{florian2006factorizing}; \cite{nadeau2007survey}; \cite{florian2010improving} in the challenging field of information extraction (IE) from biomedical text.

\par
To evaluate the efficacy of our approach, a simple baseline neural network classifier for RE, using the pretrained CharacterBERT medical version representations, is trained. The representations of the CharacterBERT tuned version after applying CL are used to train the same classifier, which vastly outperforms the baseline classifier. A tSNE \cite{van2008visualizing} analysis illustrates that meaningful relation-related clusters can be identified in the learned embedding space. This provides a second strong indication that structure can be effectively imposed on LM embeddings using our proposed framework.

\par
Even if the main focus of this work is not solving the IE problem directly, to further explore the capabilities of the relation-aware representation space, we train a simple KNN classifier for RE that is competitive with state-of-the-art performance. Strict evaluation \cite{bekoulis2018joint}; \cite{taille2020let} of the RE task presupposes correct detection of the boundaries and the entity type of each argument in the relation. Hence, we apply the CL paradigm to learn a distinct embedding space for the entities and use a KNN classifier to solve the NER task. Finally, we perform a strict evaluation of the complete entity-relation extraction task. This transparent, computationally inexpensive and intuitively simple approach has comparable results to the state-of-the-art models. This achievement illustrates how informative and meaningful the learned embedding spaces are.

\par 
In summary, our key contributions are:
\vspace{-0.5mm}
\begin{itemize}
    \item We propose a novel CL framework for imposing a relation-related
    %, which, in principle is text encoder-agnostic,
    structure on LM embeddings.
    %and leveraging the embedding space. 
    \vspace{-0.5mm}
    \item We investigate different ways to model texts and graphs and show the effectiveness of embedding relations in pairs of token embeddings.
    %a key necessity of better pooling strategies.
    \vspace{-0.5mm}
    \item 
    We exploit the capabilities of the learned representation spaces by using them in the IE task and achieve competitive results to state-of-the-art models, even if we use transparent and intuitively simple KNN classifiers. 
\end{itemize}

\par
The paper is structured as follows. Section 2 presents the ADE dataset and the data preprocessing steps, and section 3 explains the framework in detail. In section 4, we evaluate the quality of the framework in baseline setups. The tSNE analysis is presented in section 5. In section 6, we use the framework to solve the IE task and compare the results to state-of-the-art models.

\vspace{-1mm}
 
\section{Dataset}

This study focuses on biomedical text, and ADE dataset is used. The sentences are annotated with labels for drugs and adverse effects, as well as the relations among these entities. Adverse effects (AEs) cover a range of signs, symptoms, diseases, disorders, abnormalities, organ damage and even death caused by that drug. The corpus is annotated at the sentence-level, so non-local relations (between entities of different sentences) do not exist. 

\vspace{-1mm}

\subsection{Data Preprocessing}
The input of the main CL framework consists of the encoded padded sentence and the relation graph, which is extracted from the sentence. The graphs are used only in the training setup. To prepare the input for CharacterBERT, tokenization is applied to each sentence using the character-CNN module \cite{peters2018deep}. The BERT tokenizer handles out-of-vocabulary (OOV) words by splitting these words into word pieces. However, the existence of word pieces can be an obstacle in creating and testing the CL experiments of this study from the implementation point of view. Additionally, word pieces may add biases to the model \cite{el2020characterbert}, especially in biomedical text where most of the drugs and many adverse effects are OOV words. Hence, CharacterBERT is chosen instead of BERT.

\par
For each sentence, a knowledge graph is obtained to model the relations between the drugs and the adverse effects. The graph nodes are initialized with embeddings that are extracted by the final layer of the pretrained medical version of CharacterBERT. The graph convolutional network (GCN) \cite{kipf2016semi}, which is a key layer of the main proposed CL framework (Fig. 1, Fig. 2), receives two inputs: an $N$x$F$ matrix ($N$: number of nodes, $F$: number of features) with the embeddings (features) of each node and an adjacency matrix $N$x$N$, which models the connections (edges) of the undirected graph. Generally, the adjacency matrices are very sparse if we consider all the tokens and create the whole graph because the relations are rare and there are many singleton nodes. Alternatively, the tokens that are part of a relation can only be used, and the essential subgraph is extracted. For example, in the sentence "\textit{Methods: we report two cases of pseudoporphyria caused by naproxen and oxaprozin.}" There are two AE relations between AE \textit{pseudoporphyria} and the drugs \textit{naproxen} and \textit{oxaprozin}. Hence, by creating the subgraph, only these AE and drug tokens are included, and the singleton nodes (rest of the sentence tokens) are removed.

\par
The drug and the AE entities may consist of more than one word. There are two methods to model this case. On the one hand, the whole phrase can be represented as one node in the graph by averaging the embeddings of each distinct word of the phrase. On the other hand, each node refers to the last word of the entity. For example, if the initial relation is between the drug "\textit{gabapentin}" and the adverse effect "\textit{renal impairment}", then in the graph, the relation [gabapentin, impairment] is modeled. The latter approach is mainly adopted in nonspan-based relation extraction models \cite{bekoulis2018joint}; \cite{zhaomodeling}. In this study, the second approach is adopted because it gives the flexibility in applying contrastive learning at the token and relation levels.

\par 
The normalization of the adjacency matrix is essential for aggregating and propagating the information in the graph effectively \cite{kipf2016semi} and is described by the following equations:
\vspace{-2mm}
\begin{equation}
    \vspace{-2mm}
    A_{hat} = A + I,
    \vspace{-4.5mm}
\end{equation}

\begin{equation}
    A_{norm} = D^{-0.5} * A_{hat} * D^{-0.5},
\end{equation}

\noindent where $A$ is the initial adjacency matrix, $I$ is the identity matrix and $D$ is the degree matrix. 

Initially, the whole corpus is stored in one text file. Hence, the data should be transformed and stored using a different more flexible format. For each sentence of the dataset, a distinct JSON file is created and contains a list with the tokens \footnote{The sentence tokenization is performed using the SpaCy library.}, a list with NE tags adopting the BIO encoding scheme \cite{sang1999representing}; \cite{ratinov2009design}, a list with token index pairs that are members of an existing relation, the padded encoded version of the sentence, the attention mask vector of the sentence, a list with the embeddings of each node of the graph and the normalized adjacency matrix.

\vspace{-1mm}

\subsection{Dataset Statistics}

The ADE dataset is not officially split into training, validation, and test sets. Hence, we evaluate our models using 10-fold cross-validation similar to \cite{li2017neural}. We use the same splits as \cite{ebertsspan}. As \cite{taille2020let} stresses, many works on the IE task do not report the data preprocessing and detailed statistics of the datasets. This is an obstacle for a sanity check and reproducibility. The ADE dataset consists of 4,272 sentences, with 5,063 drug entities (1,048 unique drugs), 5,776 AE entities (2,983 unique AEs) and 6,821 relations. We report the statistics of each split (Table 1) and propose using this particular split for a fair comparison \footnote{To facilitate further research, the preprocessed data and the code will be publicly available in the official repository of the paper.}.

\begin{table}[!ht]
    \centering
    \resizebox{0.48\textwidth}{!}{%
        \begin{tabular}{lllll}
            \hline
            \multirow{2}{*}{\textbf{Split}} & \multicolumn{2}{c}{\textbf{Training Set}} & \multicolumn{2}{c}{\textbf{Test Set}}\\
            \phantom & {\textbf{Relation Count}} & {\textbf{Entity Count}} & {\textbf{Relation Count}} & {\textbf{Entity Count}}\\
            \hline
            1 & 6,155 & 9,769 & 666 & 1,070\\
            2 & 6,097 & 9,713 & 724 & 1,126\\
            3 & 6,133 & 9,748 & 688 & 1,091\\
            4 & 6,164 & 9,771 & 657 & 1,068\\
            5 & 6,173 & 9,785 & 648 & 1,054\\
            6 & 6,089 & 9,713 & 732 & 1,126\\
            7 & 6,155 & 9,768 & 666 & 1,071\\
            8 & 6,117 & 9,754 & 704 & 1,085\\
            9 & 6,133 & 9,760 & 688 & 1,079\\
            10 & 6,173 & 9,770 & 648 & 1,069\\
            \hline
            Mean & 6,139 & 9,755 & 682 & 1,084\\
            \hline
        \end{tabular}}
    \caption{Statistics of 10-fold splits - ADE dataset}
\end{table}

\vspace{-1mm}

\section{Framework}

In essence, contrastive learning is a paradigm for learning representations which capture some auxiliary information by training them to distinguish positive from negative instances of this auxiliary information. Our framework is inspired by the recent publications on image view-based CL of visual representation \cite{NEURIPS2020_d89a66c7}; \cite{zhang2020contrastive}; \cite{henaff2020data}; \cite{chen2020simple}; \cite{he2020momentum}, but differs from the existing work by the application of CL to the graph and text modalities. Our work is also inspired by the semantic bootstrapping hypothesis \cite{pinker1996language}, which proposes that children acquire their native language through exposure to sentences of the language (i.e., a language model) paired with structured representations of their meaning \cite{abend2017bootstrapping}.

The main CL framework for imposing relation-aware structure on the token embeddings is tested under two different settings. The difference in each setting is related to the modeling of the graph and the level of applying the CL paradigm. To solve the end-to-end IE task, a second model is proposed for learning a distinct embedding space where the named entities are projected.

\vspace{-1mm}

\subsection{Model Architectures}

In the first setting (Fig. 1), we apply the CL method to the embeddings of graph nodes in their graph context and the embeddings of sentence tokens in their sentence context.  We call this variation in the main CL framework \textit{CLGS}.  The positive and sampled negative graph representations are computed by a graph convolutional network (GCN) \cite{kipf2016semi}; \cite{schlichtkrull2018modeling} layer followed by a pooling layer.  We model the graph considering only the tokens that are part of a relation (subgraphs). To obtain one representation for the graph, average and maximum pooling strategies are tried. Tanh (range: [-1, 1]) is chosen as the activation function of the GCN layer because the text encoder also extracts negative embeddings. Hence, a similar range of embedding values should be extracted from the graph. The sentence is passed to the text encoder (CharacterBERT), which has the first six layers frozen. CharacterBERT is initialized with the pretrained weights (medical version). A pooling layer follows, to create a representation for the whole sentence. Taking the average, maximum embedding vector and the [CLS] token representation are tested as pooling strategies. The addition of a projection layers before applying CL is a common approach \cite{chen2020simple}; \cite{zhang2020contrastive}. ReLU is used as the activation function of the projection layers to introduce nonlinearity. By adding the projection layers, there is the danger that the task will be solved mainly in the projection layers, while the final goal is pushing structured relation-aware information in the text encoder. Finally, CL is applied to the resulting pair of graph and sentence representations, so that the pooled sentence token embeddings are trained to carry the information in the pooled graph node embeddings.

\vspace{-1mm}
\begin{figure}[!h]
  \centering
  \includegraphics[width=7.5cm, height=11cm, keepaspectratio]{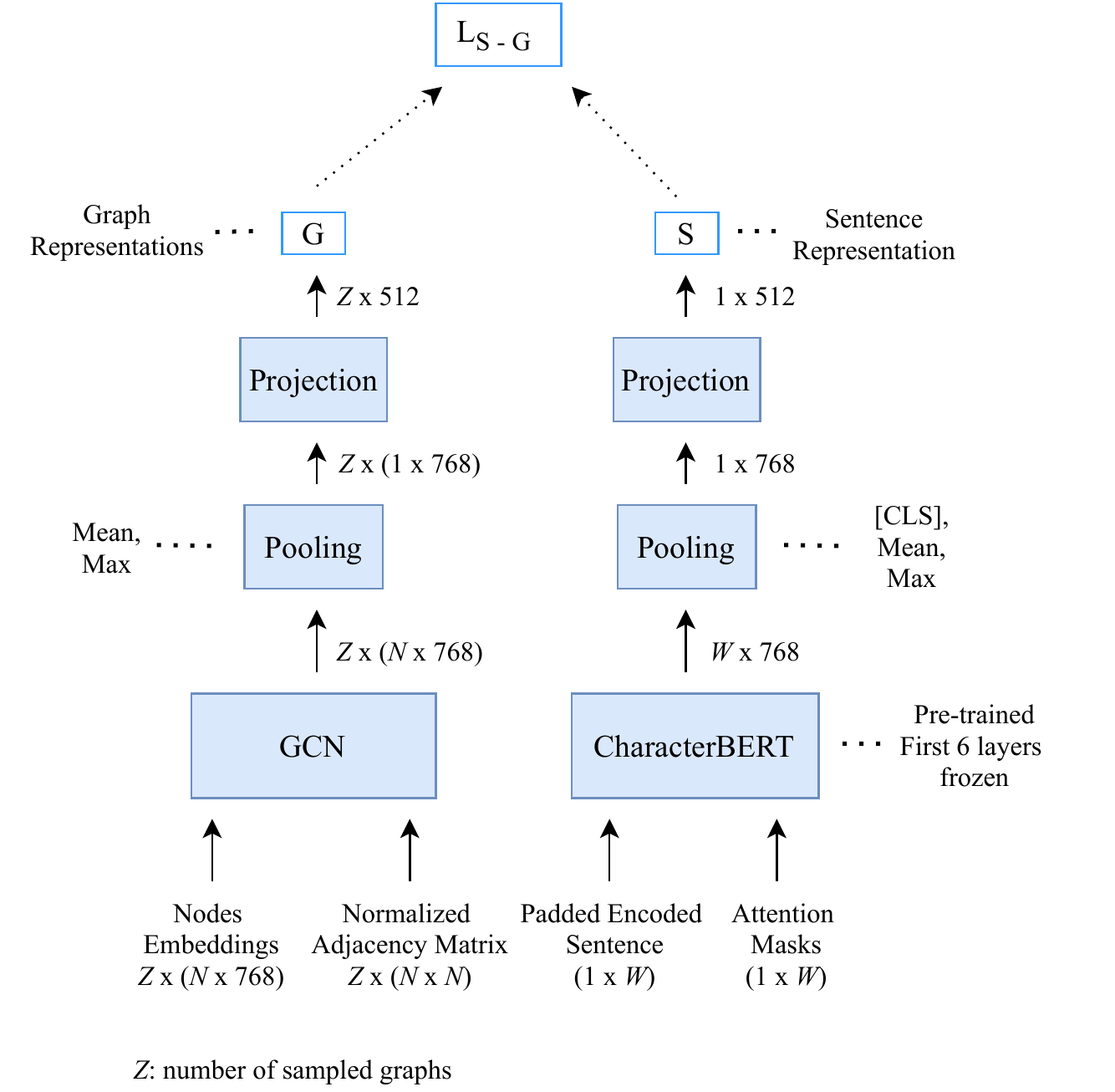}
  \vspace{-1ex}
  \caption{CL framework \textit{CLGS} - 1\textsuperscript{st} Setting}
\end{figure}
\vspace{-2mm}

In the second setting, we apply the CL method to the embeddings of graph relations and the embeddings of pairs of sentence tokens.  This variation in the CL framework is called \textit{CLDR}.  The graph is simplified to the extreme level.  Each relation is modeled completely independently in the graph, and the relation representations are extracted by concatenation of the nodes that are connected in the disjoint graphs (Fig. 2). This graph modeling makes the CL at the relation level a more tractable task. In addition, sampling negative graphs can be implemented more easily in a more controlled way. In this setting, because the graphs only have two nodes, the adjacency matrix should not be normalized in a balanced way. If the adjacency matrix is
$\big(\begin{smallmatrix}
    0.5 & 0.5\\
    0.5 & 0.5
\end{smallmatrix}\big)$
, then the final node embeddings will be the same for the two nodes that form the graph. Hence, we suggest focusing more on the self-loop of each node to keep its predefined contextualized information up to a certain level \footnote{We remind that the nodes are initialized with embeddings extracted from the pretrained CharacterBERT medical version.}. The final adjacency matrix has the following format
$\big(\begin{smallmatrix}
    \lambda & 1-\lambda\\
    1-\lambda & \lambda
\end{smallmatrix}\big)$, where \(\lambda\) is a hyperparameter of the model. The \(\lambda\) parameter defines the balance of focusing on the self-loop of each node and its neighbor (connected node). Intuitively, a \(\lambda\) value equal to 0.8 is a good choice for focusing attention on the self-loop and having distinct embeddings for the connected nodes. ReLU is used as the activation function of the GCN layer. 

On the text side, the pair of tokens that form a relation in the disjoint graphs are chosen, and the concatenation of their representations is used as the final relation representation. Finally, CL is applied on the relation level, so that the pairs of sentence token embeddings are trained to carry the information in the pairs of related graph node embeddings.

\begin{figure}[!h]
  \centering
  \includegraphics[width=7.5cm, height=11cm, keepaspectratio]{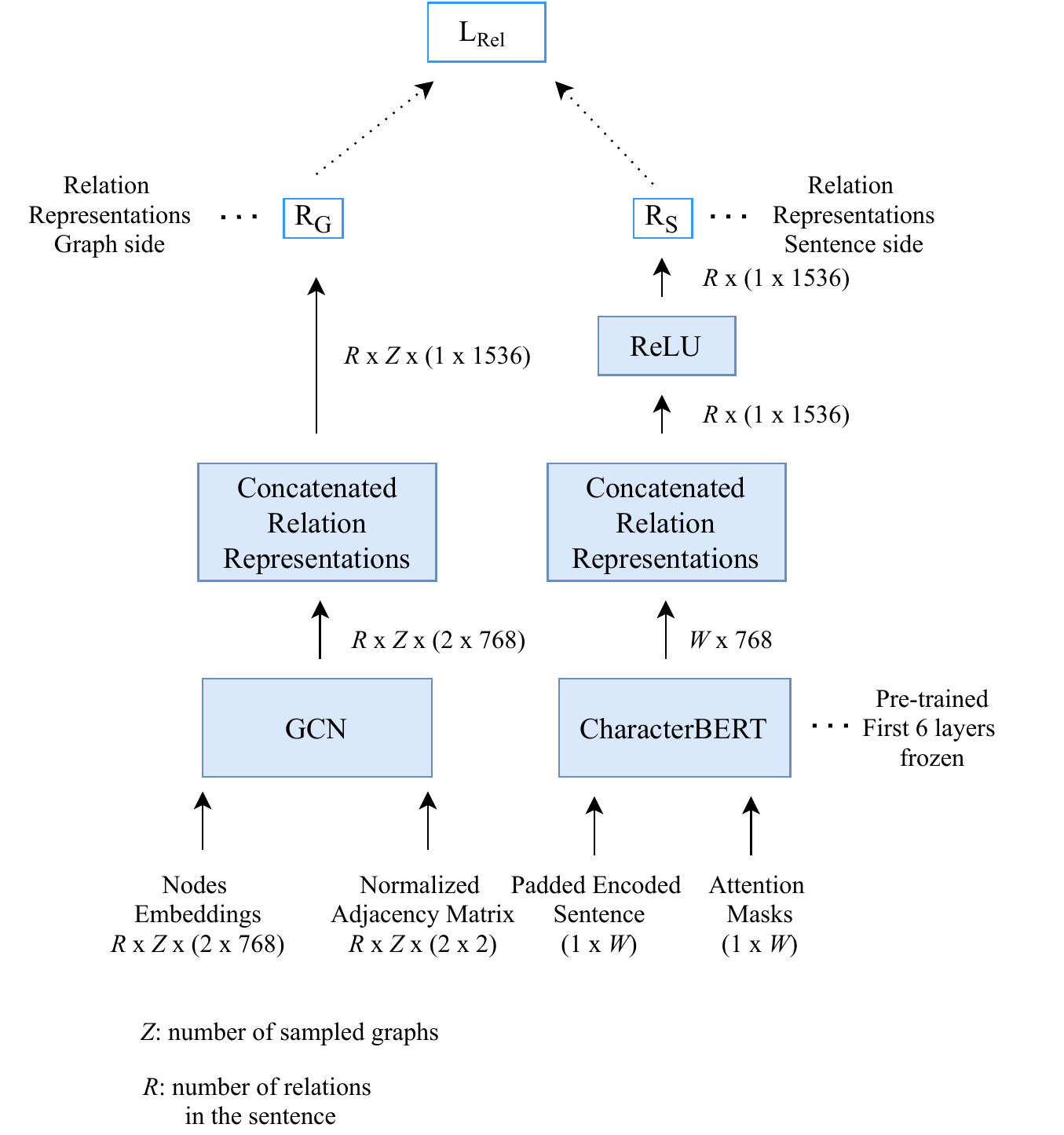}
  \vspace{-1ex}
  \caption{CL framework \textit{CLDR} - 2\textsuperscript{nd} Setting}
\end{figure}
\vspace{-2mm}

A distinct model (called \textit{CLNER}) for learning meaningful representations for named entities is designed (Fig. 3). CharacterBERT captures contextualised information very well. Hence, only one dense layer is added after CharacterBERT. Then a random sampling for the named entities is performed in a balanced way. A pool of sampled entities of the batch is selected and CL is applied on the token level.

\begin{figure}[!h]
  \centering
  \includegraphics[width=7.5cm, height=11cm, keepaspectratio]{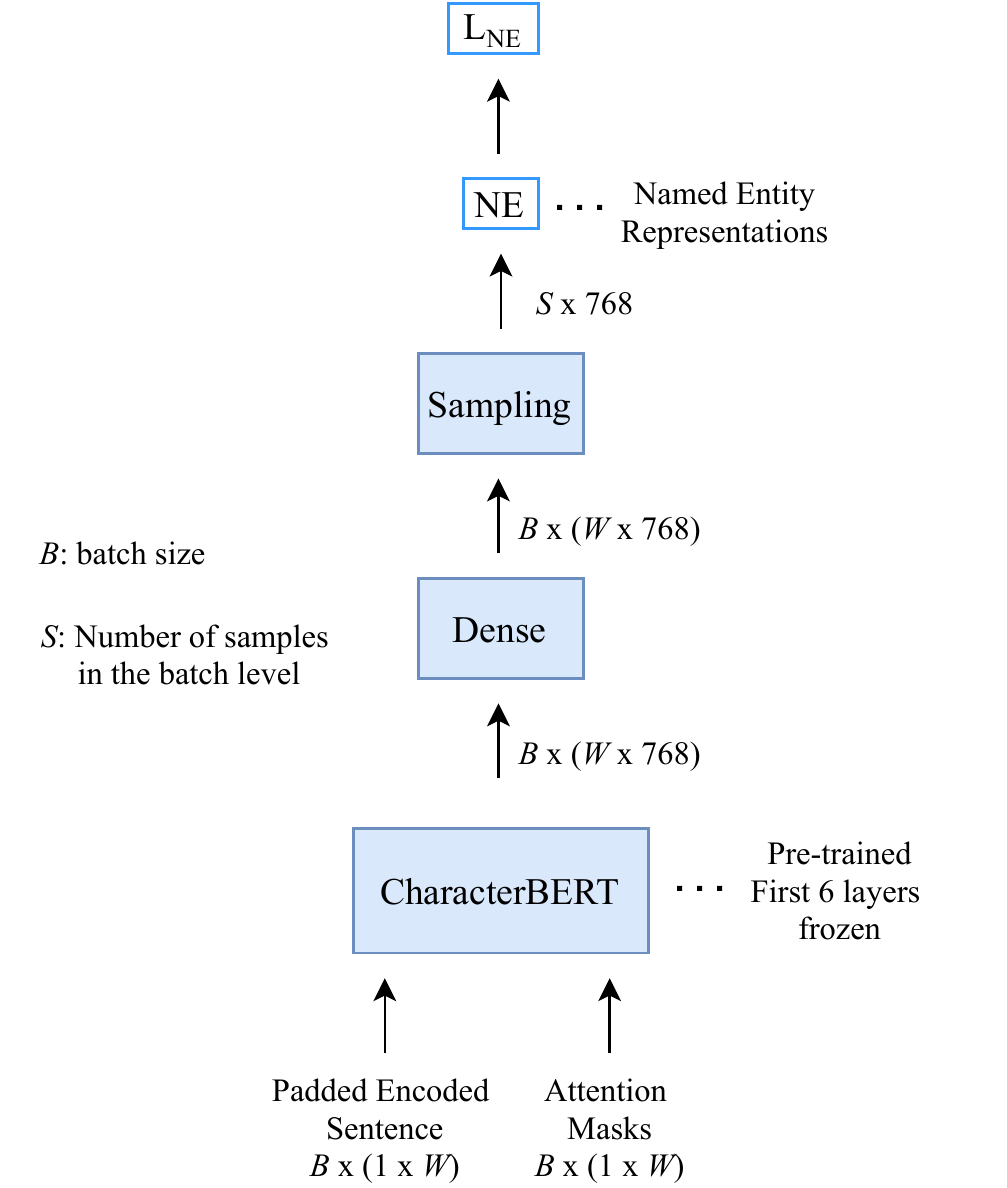}
  \vspace{-1ex}  
  \caption{Model \textit{CLNER} for learning named entity representations}
  \vspace{-4mm}
\end{figure}

\vspace{-1.5mm}

\subsection{Sampling Strategy}

Hard negative sampling is important to effectively apply the CL paradigm. The negative graphs are created by randomly selecting tokens that are not part of an adverse effect entity, keeping the correct drug tokens, and vice versa. Hence, hard incorrect drug and adverse effects relation pairs are introduced to the graph. The positive and negative graphs of each sentence have the same number of relations but not necessarily the same number of nodes. The sampling strategy is similar for the \textit{CLGS} (Fig. 4) and the \textit{CLDR} model (Fig. 5). For the \textit{CLDR} model, the positive graph is simplified to a disjoint graph, and then hard negative sampling is performed.

\begin{figure}[!h]
  \centering
  \includegraphics[width=7.5cm, height = 9cm, keepaspectratio]{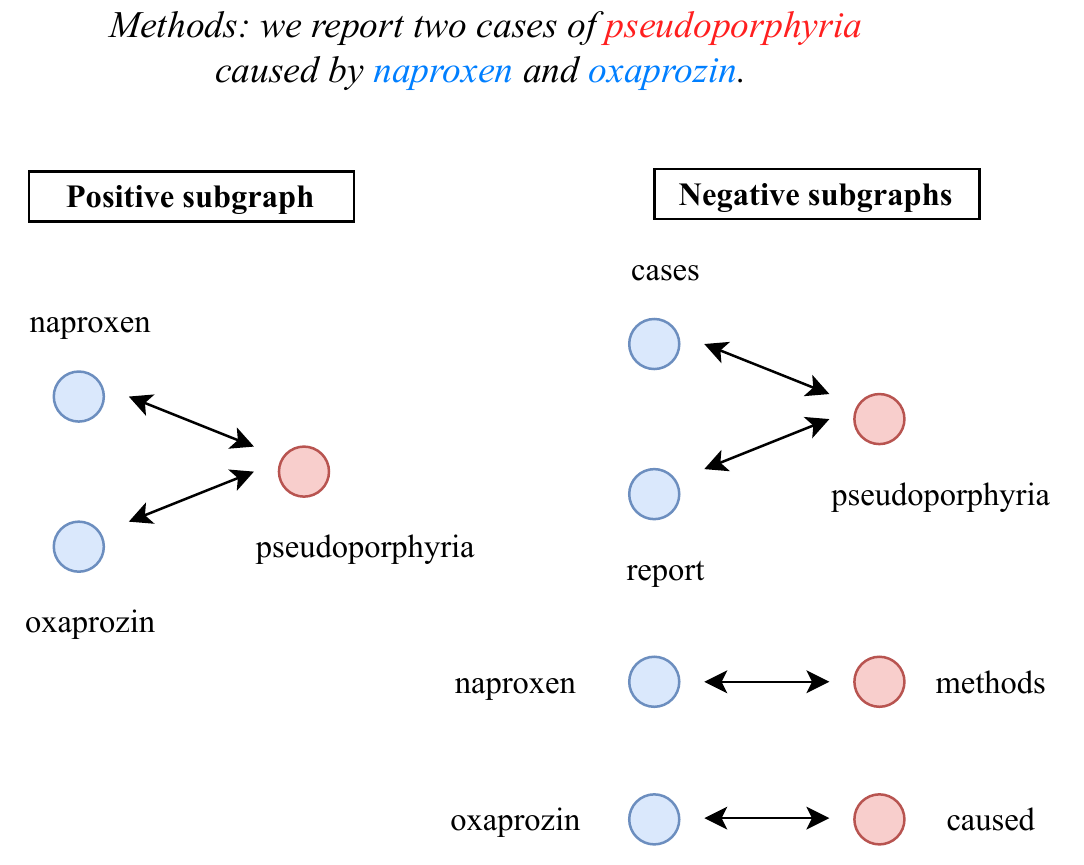}
  \vspace{-1ex} 
  \caption{Example of sampling negative graphs - \textit{CLGS} model}
\end{figure}

\begin{figure}[!h]
  \centering
  \includegraphics[width=7.5cm, height = 9cm, keepaspectratio]{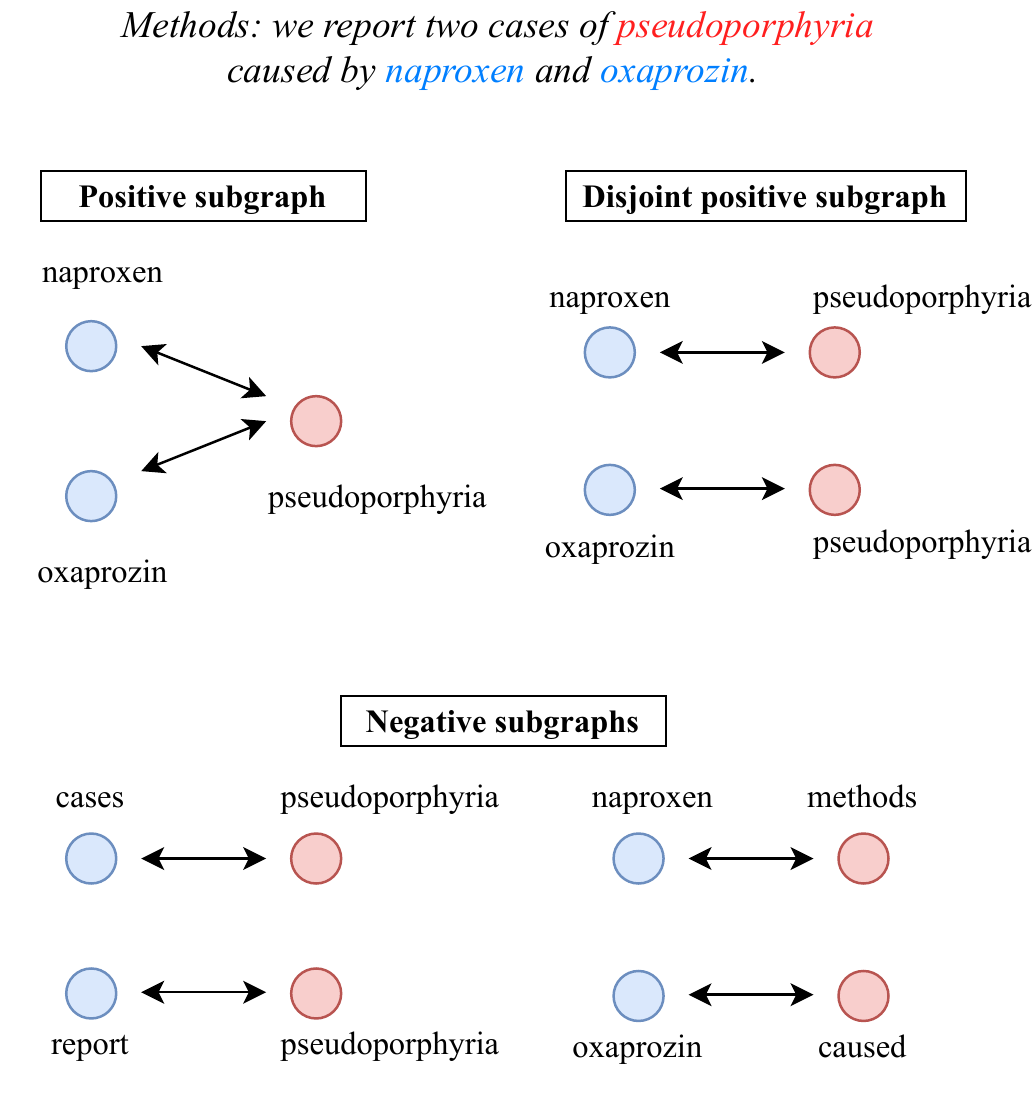}
  \vspace{-1ex} 
  \caption{Example of sampling negative graphs - \textit{CLDR} model}
  \vspace{-4.5mm}
\end{figure}

In the \textit{CLNER}, random sampling\footnote{Hard negative sampling based on the Euclidean distance and cosine similarity is also tested, but the performance is not increased. Hence, the complexity-performance trade-off leads us to finally select random sampling.} is executed at the batch level. Analysis of the number of different entity tags (drug, AE or outside token) in the batch is performed a priori to choose an appropriate number of positive and negative samples (balanced sampling).

\vspace{-1mm}

\subsection{Design Choices}

In this subsection, the justification for the model design choices is discussed. For the \textit{CLGS} and \textit{CLDR} models, the GCN layer is the key element because it can produce useful node representations considering the graph links. The propagation rule of the GCN layer is described by the following equation:

\vspace{-5.5mm} 
\begin{equation}
    X_{l+1} = \sigma(A_{norm} * X_{l} * W_{l}),
    \vspace{-0.5ex} 
\end{equation}
where \(\sigma(\cdot)\) is the activation function (e.g., ReLU, Tanh), \(A_{norm}\) is the normalized adjacency matrix (Eq. 2), \(X_{l}\) the node embeddings and \(W_{l}\) the weights of the \textit{l} layer.

In the first setting (\textit{CLGS} model), the graph is propagated through the GCN layer, and a final pooled graph representation is extracted. We hypothesize that using the CL paradigm, the model can learn which part of the information is essential for the relation representations by keeping the structure-related information in the graph representation. In the second setting (\textit{CLDR} model), the level of abstraction is reduced because instead of applying CL in the graph sentence, we use the CL paradigm at the relation level. The strategy of creating disjoint graphs results in learning similar representations for the drug and AE nodes. To address this, the relations are represented asymmetrically as a concatenation of the nodes. We hypothesize that relation-related information can be imposed in the pair-of-tokens embeddings of the LM by applying CL to them and these pair-of-nodes embeddings of the graph relations (Fig. 2).

\vspace{-1mm}

\subsection{Training Details}

The models are trained using a CL loss function that is similar to the SimCLR loss function \cite{chen2020simple}. In the first setting (\textit{CLGS} model), the main concept is to leverage the two graph and sentence representations so the true representation pair is close and similar in the learned embedding space. At each training time, a set of $Z$ graphs (the positive and some negative graphs) and the corresponding sentence are passed on the model, and the corresponding representations are calculated. Therefore, the contrastive loss receives the graph and sentence representations and for the $i$-th pair is as follows:
\vspace{-1mm}
\begin{equation}
    \resizebox{.8\hsize}{!}{$l_{i}^{(S \rightarrow G)} = -log(\frac{exp(<S_{i}, G_{i}>/\tau)}{ \sum_{z=1}^{Z} exp(<S_{i}, G_{z}>/\tau)})$},
    \vspace{-0.5ex}
\end{equation}
where \(<S_{i}, G_{i}>\) represents the cosine similarity and \(\tau\) is a temperature parameter. 

In the second setting (\textit{CLDR} model), the pair of node embeddings that are extracted from the disjoint graphs encode their relation, because this is the main functionality of the GCN layer.  Hence, the main idea is to increase the similarity between the representations of the correct relation in the graph and the relation representations that are extracted from the text encoder.
%and decrease the similarity between incorrect pairs. 
The contrastive loss for each sentence is as follows:
\vspace{-1mm}
\begin{equation}
    \resizebox{.86\hsize}{!}{$l^{(RS \rightarrow RG)} = \sum_{r=1}^{R}-log(\frac{exp(<RS_{r}, RG_{r}>/\tau)}{ \sum_{z=1}^{Z} exp(<RS_{r}, RG_{z}>/\tau)})$},
    \vspace{-0.5ex}
\end{equation}
where $R$ is the total number of relations in the sentence, $RS$ is the relation representation of the text encoder and $RG$ is the relation representation of the graph. 

\newpage
For the \textit{CLNER} model, the contrastive loss is as follows:
\vspace{-1mm}
\begin{equation}
    \resizebox{.86\hsize}{!}{$l_{NE} = \sum_{n=1}^{N}-log(\frac{\sum_{p=1}^{P} exp(<RN_{n}, RN_{p}>/\tau)}{ \sum_{k=1}^{K} exp(<RN_{n}, RN_{k}>/\tau)})$},
    \vspace{-0.5ex}
\end{equation}
where $N$ is the total number of tokens in the batch, $P$ is the number of the positive samples (same NE tag), $K$ is the total number of samples and $RN$ is the extracted token representation. 

We use a batch-size of 8 for training the \textit{CLGS} and \textit{CLDR} models, and 16 for the \textit{CLNER} model. ADAM optimizer \cite{kingma2014adam} is selected with a learning rate of 1e-5 \footnote{More information about hyperparameter tuning-selection is given in the Appendix section.}.

\vspace{-1mm}

\section{Evaluation - Baseline}

For the \textit{CLGS} model, the first evaluation step is a simple similarity check. We use the trained \textit{CLGS} model to extract the sentence representation and the positive and negative graph representations for all the sentences in the test set. Then, a similarity check is applied using the extracted sentence and graph representations. The most similar graph is predicted as the positive sentence graph. Given the positive and all the negative hard graphs extracted from each sentence, the model should be able to detect the correct graph. The different model variations perform well, but the mean pooling selection in the graph and sentence side results in better performance, as the accuracy is over 91\%. The addition of the projection layers is not advantageous.

\begin{table}[!ht]
    \centering
    \resizebox{0.48\textwidth}{!}{%
        \begin{tabular}{llll}
            \hline
            \textbf{Graph Pooling} & \textbf{Text Pooling} & \textbf{Projection layer} & {\textbf{Accuracy}}\\
            \hline
            Mean & [CLS] & - & 88.39\\
            Mean & Mean & - & \textbf{91.23}\\
            Max & Max & - & 89.1\\
            Mean & [CLS] & Yes & 88.63\\
            Mean & Mean & Yes & 87.68\\
            \hline
        \end{tabular}}
    \caption{Results - \textit{CLGS} model: Finding the correct graph with similarity check}
\end{table}

\vspace{-3mm}
The second evaluation step is applied to both models (\textit{CLGS} and \textit{CLDR}). Following previous research on representation learning \cite{henaff2020data}; \cite{chen2020simple}; \cite{he2020momentum}; \cite{zhang2020contrastive}, we evaluate the tuned CharacterBERT text encoder, taken from the trained \textit{CLGS} and \textit{CLDR} models, in a linear classification setting, where all the candidate relations (concatenation of the token embeddings) are created, and a linear classification layer is trained for the RE task. As a baseline model, we use the pretrained medical CharacterBERT to create the representation for the relations\footnote{We also try fine-tuning both the text encoder and the linear head, but the performance is not improved.}. This linear setting directly provides insight into how successfully the relation-related structure is imposed at the token level of the text encoder, by evaluating the quality of the learned representations for RE.

\begin{table}[!ht]
    \centering
    \resizebox{0.48\textwidth}{!}{%
        \begin{tabular}{llll}
            \hline
            \textbf{Model} & \textbf{Precision} & \textbf{Recall} & {\textbf{F1}}\\
            \hline
            Baseline & 69.96 & 64.39 & 66.79\\
            CharacterBERT\textsubscript{CLGS} & 56.82 & 59.42 & 58.09\\
            CharacterBERT\textsubscript{CLDR} & \textbf{79.51} & \textbf{84.39} & \textbf{81.73}\\
            \hline
        \end{tabular}}
    \caption{RE - linear classification setting}
\end{table}

 \vspace{-3mm}
Using the tuned CharacterBERT representation from the \textit{CLGS} model (mean graph and text pooling) results in poor performance. The pooling layer smooths the information. Hence, structure-related information cannot be passed at the token level of the text encoder. A smarter pooling strategy that preserves most of the relation-aware information would be ideal, but designing such pooling is difficult. The main obstacle is the varied number of relations. In contrast, when we use the tuned CharacterBERT of the \textit{CLDR} model, the basic classifier vastly outperforms the baseline model. This is a strong indication that the relation-related structure is successfully imposed on the pairs of token embeddings of the text encoder.

\vspace{-1mm}
\section{tSNE Analysis}

A tSNE analysis is performed to further explore the quality of the learned embedding spaces. Using the tuned CharacterBERT of the \textit{CLDR} model, the relation representation space is created. We project the positive (orange dots) and hard negative relations (blue dots), where one of the two relation tokens is correct. In the tSNE plot (Fig. 6), meaningful relation clusters can be easily identified, which demonstrates the efficiency of our framework (\textit{CLDR} model). The relation representations are asymmetric, as the drug and AE tokens have similar representations (Fig. 7). This means that we cannot solve RE and NER tasks using the same representation space. Hence, we learn a different space for the named entities (\textit{CLNER} model).

\par
In the tSNE plot in the entity representation space (Fig. 8), we can detect insightful entity clusters. In particular, the clusters related to the drug tags (B-DRUG, I-DRUG) are very dense and well shaped. This is a strong finding that illustrates that the \textit{CLNER} model can extract very good representations for the NER task.

\begin{figure}[!h]
  \centering
  \includegraphics[width=7.5cm, height = 9cm, keepaspectratio]{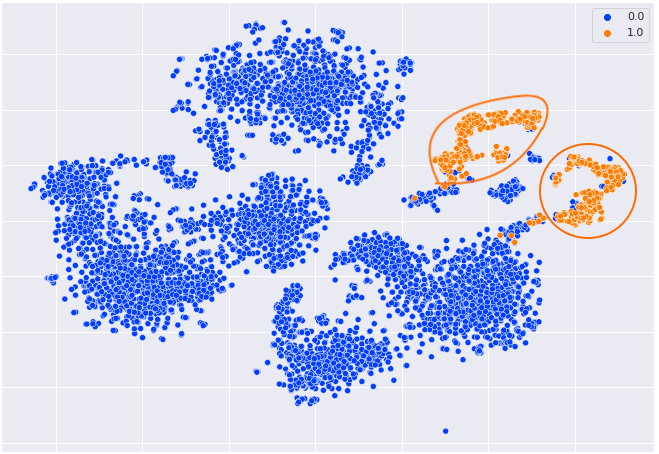}
  \vspace{-1.5ex} 
  \caption{tSNE plot - Relation representation space obtained with CharacterBERT of \textit{CLDR} model (1: relation, 0: no relation)}
  \vspace{-4mm}
\end{figure}

\begin{figure}[!h]
  \centering
  \includegraphics[width=7.5cm, height = 9cm, keepaspectratio]{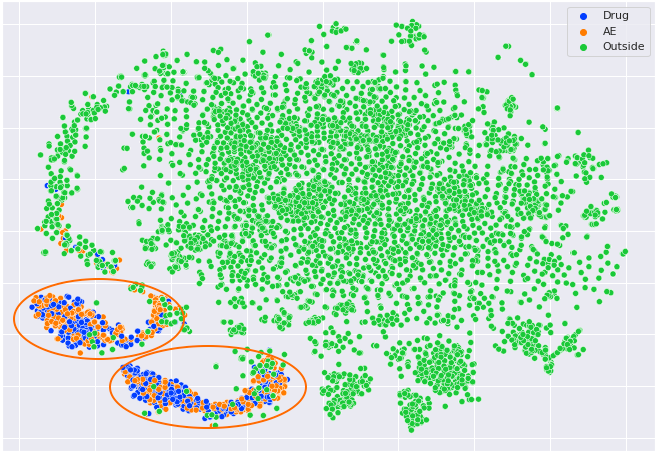}
  \vspace{-1.5ex} 
  \caption{tSNE plot - Relation representation space obtained with CharacterBERT of \textit{CLDR} model - Named Entities}
  \vspace{-4mm}
\end{figure}

\begin{figure}[!h]
  \centering
  \includegraphics[width=7.5cm, height = 9cm, keepaspectratio]{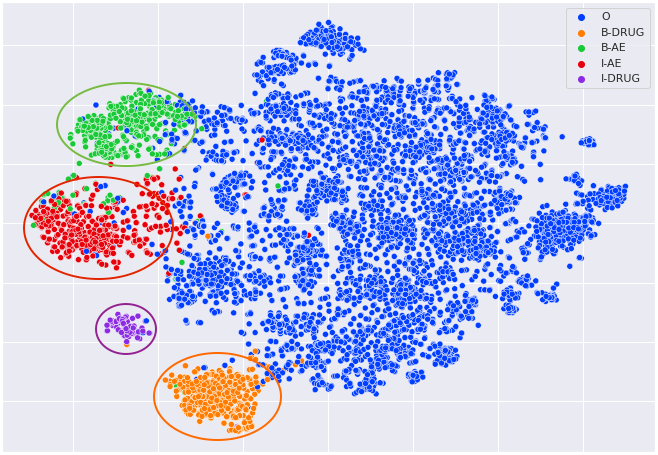}
  \vspace{-1.5ex} 
  \caption{tSNE plot - Entity representation space obtained with \textit{CLNER} model}
  \vspace{-2mm}
\end{figure}

\vspace{-4mm}
\section{Entity-Relation task}

The insights of the tSNE analysis, with the well-defined clusters in the embedding spaces, lead us to approach the entity-relation task using intuitively simple and transparent KNN classifiers. For the RE task, we utilize the tuned CharacterBERT of the \textit{CLDR} model to create the candidate relation representations. At the inference step, for each candidate relation, we decide whether it is positive based on the labels of the $k$-nearest neighbors in the learned embedding space. The value of $k$ is chosen based on the performance in the randomly selected validation set (10\% of training set) for each fold. We adopt the same strategy for the NER task using the \textit{CLNER} model and project each token to the named entity representation space.

\par
To solve both NER and RE tasks, we combine the two semantic spaces. First, we determine whether a candidate relation (concatenation of the tokens) is predicted as positive in the relation representation space, which is obtained by the tuned CharacterBERT of the \textit{CLDR} model. Then, we determine whether the boundaries and the types of the two entities in the candidate relation are predicted correctly in the entity representation space obtained by the \textit{CLNER} model. All possible candidate relations and the named entities of the test set are classified.

\par
We strictly evaluate the performance of the IE task. As \cite{bekoulis2018joint} state, an entity is considered correct if its boundaries are detected correctly and the predicted type (drug or AE) matches the ground truth. In the same setup, a relation is considered correct if its type and the two entities (boundaries and type) involved in the relation are correctly predicted. We measure precision, recall and F1 score. Following previous work on IE, we report the macro-averaged F1 score, and as 10-fold cross-validation is adopted, we average the scores over the folds.

\begin{table}[!ht]
    \centering
    \resizebox{0.48\textwidth}{!}{%
        \begin{tabular}{llll}
            \hline
            \textbf{Model} & \textbf{NER} & \textbf{RE} & {\textbf{RE-}}\\
            \hline
            \citealp{li2016joint} & 79.5 & 63.4 & -\\
            \citealp{li2017neural} & 84.6 & 71.4 & -\\
            \citealp{bekoulis2018joint} & 86.4 & 74.58 & -\\
            \citealp{bekoulis2018adversarial} & 86.73 & 75.52 & -\\
            \citealp{tran2019neural} & 87.11 & 77.29 & -\\
            \citealp{ebertsspan} & 89.25 & 79.24 & -\\
            \citealp{wang2020two} & 89.7 & 80.1 & -\\
            \citealp{zhaomodeling} & 89.4 & 81.14 & -\\
            \hline
            Ours & 88.3 & 79.97 & 86.5\\
            \hline 
        \end{tabular}}
        \vspace{-1ex}
    \caption{Test set results: macro-averaged F1 score}
    \vspace{-4mm}
\end{table}

\par
In the RE task, we achieve very competitive results using a simple and transparent KNN classifier. In contrast, the state-of-the-art models \cite{wang2020two}; \cite{zhaomodeling} are very complex and computationally expensive. This fact highlights the high quality of the learned relation representation space (\textit{CLDR} model). In principle, the NER task is a sequence-tagging problem. However, we obtain good performance with a KNN classifier that performs the inference in the learned entity representation space (\textit{CLNER} model).

\par
Notably, the last column of Table 4 (\textit{RE-}) presents the performance of the RE KNN classifier in predicting whether there is a relation between two tokens, without considering the NER task (type and boundaries of the entities). In this case, the F1 score is 86.5, and this value is the upper bound performance of our approach. Hence, incorporating a state-of-the-art model for the NER task (e.g., Wang and Lu, 2020, Eberts and Ulges, 2020) could further improve the scores of the RE task under strict evaluation. However, we use the SpERT model \cite{ebertsspan} for NER (F1 score: 89.25), but the results in the RE task are not improved. This illustrates that our NER results are already very competitive.

\par 
The above results reveal the quality of the representations for both NER and RE tasks. Hence, the proposed CL framework can be used as a preprocessing and representation learning step in the pipeline for IE models. The CL framework can be trained to leverage the embedding space and create meaningful, disentangled representations for the IE task. We successfully evaluated the representations with a simple KNN classifier, but the learned representations can be used as input in complex models for entity and relation classification to achieve better results and faster convergence. We will explore this research direction in the future.

\vspace{-1mm}
\section{Conclusion}
\vspace{-0.5mm}
We present a novel CL framework, which, in principle, is text encoder-agnostic, for effectively imposing relation-related structure to LMs and leveraging the embedding space. We evaluate the quality of the learned representations using relative baselines and competitively solve an entity-relation task. The overall results indicate that the learned representations are very powerful. The performed tSNE analysis illustrates that meaningful clusters can be easily identified in the learned embedding spaces. We note that the proposed framework can be used as a representation learning step for complex IE systems. In future work, we intend to explore the capabilities of our approach in continual learning settings and exploit external graph structured knowledge in representation learning of language data.

% add to camera ready:
%\section*{Acknowledgments}
%
%This work is supported by the Research Foundation – Flanders (FWO).
%...

\bibliography{CoNLL2021}

\begin{thebibliography}{37}
\expandafter\ifx\csname natexlab\endcsname\relax\def\natexlab#1{#1}\fi

\bibitem[{Abend et~al.(2017)Abend, Kwiatkowski, Smith, Goldwater, and
  Steedman}]{abend2017bootstrapping}
Omri Abend, Tom Kwiatkowski, Nathaniel~J Smith, Sharon Goldwater, and Mark
  Steedman. 2017.
\newblock \href {https://doi.org/10.1016/j.cognition.2017.02.009}
  {Bootstrapping language acquisition}.
\newblock \emph{Cognition}, 164:116--143.

\bibitem[{Bekoulis et~al.(2018{\natexlab{a}})Bekoulis, Deleu, Demeester, and
  Develder}]{bekoulis2018adversarial}
Giannis Bekoulis, Johannes Deleu, Thomas Demeester, and Chris Develder.
  2018{\natexlab{a}}.
\newblock \href {https://doi.org/10.18653/v1/D18-1307} {Adversarial training
  for multi-context joint entity and relation extraction}.
\newblock In \emph{Proceedings of the 2018 Conference on Empirical Methods in
  Natural Language Processing}, pages 2830--2836.

\bibitem[{Bekoulis et~al.(2018{\natexlab{b}})Bekoulis, Deleu, Demeester, and
  Develder}]{bekoulis2018joint}
Giannis Bekoulis, Johannes Deleu, Thomas Demeester, and Chris Develder.
  2018{\natexlab{b}}.
\newblock \href {https://doi.org/https://doi.org/10.1016/j.eswa.2018.07.032}
  {Joint entity recognition and relation extraction as a multi-head selection
  problem}.
\newblock \emph{Expert Systems with Applications}, 114:34--45.

\bibitem[{Brown et~al.(2020)Brown, Mann, Ryder, Subbiah, Kaplan, Dhariwal,
  Neelakantan, Shyam, Sastry, Askell, Agarwal, Herbert-Voss, Krueger, Henighan,
  Child, Ramesh, Ziegler, Wu, Winter, Hesse, Chen, Sigler, Litwin, Gray, Chess,
  Clark, Berner, McCandlish, Radford, Sutskever, and
  Amodei}]{NEURIPS2020_1457c0d6}
Tom Brown, Benjamin Mann, Nick Ryder, Melanie Subbiah, Jared~D Kaplan, Prafulla
  Dhariwal, Arvind Neelakantan, Pranav Shyam, Girish Sastry, Amanda Askell,
  Sandhini Agarwal, Ariel Herbert-Voss, Gretchen Krueger, Tom Henighan, Rewon
  Child, Aditya Ramesh, Daniel Ziegler, Jeffrey Wu, Clemens Winter, Chris
  Hesse, Mark Chen, Eric Sigler, Mateusz Litwin, Scott Gray, Benjamin Chess,
  Jack Clark, Christopher Berner, Sam McCandlish, Alec Radford, Ilya Sutskever,
  and Dario Amodei. 2020.
\newblock \href
  {https://proceedings.neurips.cc/paper/2020/file/1457c0d6bfcb4967418bfb8ac142f64a-Paper.pdf}
  {Language models are few-shot learners}.
\newblock In \emph{Advances in Neural Information Processing Systems},
  volume~33, pages 1877--1901. Curran Associates, Inc.

\bibitem[{Chen et~al.(2020)Chen, Kornblith, Norouzi, and
  Hinton}]{chen2020simple}
Ting Chen, Simon Kornblith, Mohammad Norouzi, and Geoffrey Hinton. 2020.
\newblock \href {http://proceedings.mlr.press/v119/chen20j.html} {A simple
  framework for contrastive learning of visual representations}.
\newblock In \emph{International Conference on Machine Learning}, pages
  1597--1607. PMLR.

\bibitem[{Curran and Clark(2003)}]{curran2003language}
James~R Curran and Stephen Clark. 2003.
\newblock \href {https://doi.org/https://doi.org/10.3115/1119176.1119200}
  {Language independent ner using a maximum entropy tagger}.
\newblock In \emph{Proceedings of the Seventh Conference on Natural language
  Learning at HLT-NAACL 2003}, pages 164--167.

\bibitem[{Devlin et~al.(2018)Devlin, Chang, Lee, and
  Toutanova}]{devlin2018bert}
Jacob Devlin, Ming-Wei Chang, Kenton Lee, and Kristina Toutanova. 2018.
\newblock \href {https://arxiv.org/abs/1810.04805} {{BERT}: Pre-training of
  deep bidirectional transformers for language understanding}.
\newblock \emph{arXiv preprint arXiv:1810.04805}.

\bibitem[{Eberts and Ulges(2020)}]{ebertsspan}
Markus Eberts and Adrian Ulges. 2020.
\newblock \href {https://doi.org/10.3233/FAIA200321} {Span-based joint entity
  and relation extraction with transformer pre-training}.
\newblock pages 2006--2013.

\bibitem[{El~Boukkouri et~al.(2020)El~Boukkouri, Ferret, Lavergne, Noji,
  Zweigenbaum, and Tsujii}]{el2020characterbert}
Hicham El~Boukkouri, Olivier Ferret, Thomas Lavergne, Hiroshi Noji, Pierre
  Zweigenbaum, and Jun’ichi Tsujii. 2020.
\newblock \href {https://doi.org/10.18653/v1/2020.coling-main.609}
  {Characterbert: Reconciling elmo and bert for word-level open-vocabulary
  representations from characters}.
\newblock In \emph{Proceedings of the 28th International Conference on
  Computational Linguistics}, pages 6903--6915.

\bibitem[{Florian et~al.(2006)Florian, Jing, Kambhatla, and
  Zitouni}]{florian2006factorizing}
Radu Florian, Hongyan Jing, Nanda Kambhatla, and Imed Zitouni. 2006.
\newblock \href {https://doi.org/https://doi.org/10.3115/1220175.1220235}
  {Factorizing complex models: a case study in mention detection}.
\newblock In \emph{Proceedings of the 21st International Conference on
  Computational Linguistics and the 44th Annual Meeting of the Association for
  Computational Linguistics}, pages 473--480.

\bibitem[{Florian et~al.(2010)Florian, Pitrelli, Roukos, and
  Zitouni}]{florian2010improving}
Radu Florian, John~F Pitrelli, Salim Roukos, and Imed Zitouni. 2010.
\newblock \href {https://www.aclweb.org/anthology/D10-1033} {Improving mention
  detection robustness to noisy input}.
\newblock In \emph{Proceedings of the 2010 Conference on Empirical Methods in
  Natural Language Processing}, pages 335--345.

\bibitem[{Gurulingappa et~al.(2012)Gurulingappa, Rajput, Roberts, Fluck,
  Hofmann-Apitius, and Toldo}]{gurulingappa2012development}
Harsha Gurulingappa, Abdul~Mateen Rajput, Angus Roberts, Juliane Fluck, Martin
  Hofmann-Apitius, and Luca Toldo. 2012.
\newblock \href {https://doi.org/https://doi.org/10.1016/j.jbi.2012.04.008}
  {Development of a benchmark corpus to support the automatic extraction of
  drug-related adverse effects from medical case reports}.
\newblock \emph{Journal of Biomedical Informatics}, 45(5):885--892.

\bibitem[{He et~al.(2020)He, Fan, Wu, Xie, and Girshick}]{he2020momentum}
Kaiming He, Haoqi Fan, Yuxin Wu, Saining Xie, and Ross Girshick. 2020.
\newblock \href
  {https://openaccess.thecvf.com/content_CVPR_2020/html/He_Momentum_Contrast_for_Unsupervised_Visual_Representation_Learning_CVPR_2020_paper.html}
  {Momentum contrast for unsupervised visual representation learning}.
\newblock In \emph{Proceedings of the IEEE/CVF Conference on Computer Vision
  and Pattern Recognition}, pages 9729--9738.

\bibitem[{Henaff(2020)}]{henaff2020data}
Olivier Henaff. 2020.
\newblock \href {http://proceedings.mlr.press/v119/henaff20a.html}
  {Data-efficient image recognition with contrastive predictive coding}.
\newblock In \emph{International Conference on Machine Learning}, pages
  4182--4192. PMLR.

\bibitem[{Jiang and Zhai(2007)}]{jiang2007systematic}
Jing Jiang and ChengXiang Zhai. 2007.
\newblock \href {https://www.aclweb.org/anthology/N07-1015} {A systematic
  exploration of the feature space for relation extraction}.
\newblock In \emph{Human Language Technologies 2007: The Conference of the
  North American Chapter of the Association for Computational Linguistics;
  Proceedings of the Main Conference}, pages 113--120.

\bibitem[{Khosla et~al.(2020)Khosla, Teterwak, Wang, Sarna, Tian, Isola,
  Maschinot, Liu, and Krishnan}]{NEURIPS2020_d89a66c7}
Prannay Khosla, Piotr Teterwak, Chen Wang, Aaron Sarna, Yonglong Tian, Phillip
  Isola, Aaron Maschinot, Ce~Liu, and Dilip Krishnan. 2020.
\newblock \href
  {https://proceedings.neurips.cc/paper/2020/file/d89a66c7c80a29b1bdbab0f2a1a94af8-Paper.pdf}
  {Supervised contrastive learning}.
\newblock In \emph{Advances in Neural Information Processing Systems},
  volume~33, pages 18661--18673. Curran Associates, Inc.

\bibitem[{Kingma and Ba(2014)}]{kingma2014adam}
Diederik~P Kingma and Jimmy Ba. 2014.
\newblock \href {https://arxiv.org/abs/1412.6980} {Adam: A method for
  stochastic optimization}.
\newblock \emph{arXiv preprint arXiv:1412.6980}.

\bibitem[{Kipf and Welling(2016)}]{kipf2016semi}
Thomas~N Kipf and Max Welling. 2016.
\newblock \href {https://arxiv.org/abs/1609.02907} {Semi-supervised
  classification with graph convolutional networks}.
\newblock \emph{arXiv preprint arXiv:1609.02907}.

\bibitem[{Li et~al.(2017)Li, Zhang, Fu, and Ji}]{li2017neural}
Fei Li, Meishan Zhang, Guohong Fu, and Donghong Ji. 2017.
\newblock \href {https://doi.org/10.1186/s12859-017-1609-9} {A neural joint
  model for entity and relation extraction from biomedical text}.
\newblock \emph{BMC bioinformatics}, 18(1):1--11.

\bibitem[{Li et~al.(2016)Li, Zhang, Zhang, and Ji}]{li2016joint}
Fei Li, Yue Zhang, Meishan Zhang, and Donghong Ji. 2016.
\newblock \href {https://www.ijcai.org/Abstract/16/403} {Joint models for
  extracting adverse drug events from biomedical text}.
\newblock In \emph{Proceedings of the Twenty-Fifth International Joint
  Conference on Artificial Intelligence}, IJCAI'16, page 2838–2844. AAAI
  Press.

\bibitem[{Liu et~al.(2019)Liu, Ott, Goyal, Du, Joshi, Chen, Levy, Lewis,
  Zettlemoyer, and Stoyanov}]{liu2019roberta}
Yinhan Liu, Myle Ott, Naman Goyal, Jingfei Du, Mandar Joshi, Danqi Chen, Omer
  Levy, Mike Lewis, Luke Zettlemoyer, and Veselin Stoyanov. 2019.
\newblock \href {https://arxiv.org/abs/1907.11692} {Roberta: A robustly
  optimized {BERT} pretraining approach}.
\newblock \emph{arXiv preprint arXiv:1907.11692}.

\bibitem[{Nadeau and Sekine(2007)}]{nadeau2007survey}
David Nadeau and Satoshi Sekine. 2007.
\newblock \href {https://doi.org/https://doi.org/10.1075/li.30.1.03nad} {A
  survey of named entity recognition and classification}.
\newblock \emph{Lingvisticae Investigationes}, 30(1):3--26.

\bibitem[{Peters et~al.(2018)Peters, Neumann, Iyyer, Gardner, Clark, Lee, and
  Zettlemoyer}]{peters2018deep}
Matthew~E Peters, Mark Neumann, Mohit Iyyer, Matt Gardner, Christopher Clark,
  Kenton Lee, and Luke Zettlemoyer. 2018.
\newblock \href {https://arxiv.org/abs/1802.05365} {Deep contextualized word
  representations}.
\newblock \emph{arXiv preprint arXiv:1802.05365}.

\bibitem[{Pinker(1996)}]{pinker1996language}
Steven Pinker. 1996.
\newblock \emph{Language Learnability and Language Development: With New
  Commentary by the Author}, volume~7.
\newblock Harvard University Press.

\bibitem[{Plank and Moschitti(2013)}]{plank2013embedding}
Barbara Plank and Alessandro Moschitti. 2013.
\newblock \href {https://www.aclweb.org/anthology/P13-1147} {Embedding semantic
  similarity in tree kernels for domain adaptation of relation extraction}.
\newblock In \emph{Proceedings of the 51st Annual Meeting of the Association
  for Computational Linguistics (Volume 1: Long Papers)}, pages 1498--1507.

\bibitem[{Ratinov and Roth(2009)}]{ratinov2009design}
Lev Ratinov and Dan Roth. 2009.
\newblock \href {https://www.aclweb.org/anthology/W09-1119.pdf} {Design
  challenges and misconceptions in named entity recognition}.
\newblock In \emph{Proceedings of the Thirteenth Conference on Computational
  Natural Language Learning (CoNLL-2009)}, pages 147--155.

\bibitem[{Sang and Veenstra(1999)}]{sang1999representing}
Erik F Tjong~Kim Sang and Jorn Veenstra. 1999.
\newblock \href {https://doi.org/https://doi.org/10.3115/977035.977059}
  {Representing text chunks}.
\newblock In \emph{Proceedings of the Ninth Conference of the European Chapter
  of the Association for Computational Linguistics}, pages 173--179.

\bibitem[{Schlichtkrull et~al.(2018)Schlichtkrull, Kipf, Bloem, Van Den~Berg,
  Titov, and Welling}]{schlichtkrull2018modeling}
Michael Schlichtkrull, Thomas~N Kipf, Peter Bloem, Rianne Van Den~Berg, Ivan
  Titov, and Max Welling. 2018.
\newblock \href {https://doi.org/https://doi.org/10.1007/978-3-319-93417-4_38}
  {Modeling relational data with graph convolutional networks}.
\newblock In \emph{European Semantic Web Conference}, pages 593--607. Springer.

\bibitem[{Sun et~al.(2011)Sun, Grishman, and Sekine}]{sun2011semi}
Ang Sun, Ralph Grishman, and Satoshi Sekine. 2011.
\newblock \href {https://www.aclweb.org/anthology/P11-1053} {Semi-supervised
  relation extraction with large-scale word clustering}.
\newblock In \emph{Proceedings of the 49th Annual Meeting of the Association
  for Computational Linguistics: Human Language Technologies}, pages 521--529.

\bibitem[{Taill{\'e} et~al.(2020)Taill{\'e}, Guigue, Scoutheeten, and
  Gallinari}]{taille2020let}
Bruno Taill{\'e}, Vincent Guigue, Geoffrey Scoutheeten, and Patrick Gallinari.
  2020.
\newblock \href {https://www.aclweb.org/anthology/2020.emnlp-main.301.pdf}
  {Let’s stop error propagation in the end-to-end relation extraction
  literature!}
\newblock In \emph{Proceedings of the 2020 Conference on Empirical Methods in
  Natural Language Processing ({EMNLP} 2020)}, pages 3689--3701.

\bibitem[{Tran and Kavuluru(2019)}]{tran2019neural}
Tung Tran and Ramakanth Kavuluru. 2019.
\newblock \href {https://arxiv.org/abs/1905.07458} {Neural metric learning for
  fast end-to-end relation extraction}.
\newblock \emph{arXiv preprint arXiv:1905.07458}.

\bibitem[{Van~der Maaten and Hinton(2008)}]{van2008visualizing}
Laurens Van~der Maaten and Geoffrey Hinton. 2008.
\newblock \href {https://www.jmlr.org/papers/v9/vandermaaten08a.html}
  {Visualizing data using t-sne.}
\newblock \emph{Journal of Machine Learning Research}, 9(11).

\bibitem[{Vaswani et~al.(2017)Vaswani, Shazeer, Parmar, Uszkoreit, Jones,
  Gomez, Kaiser, and Polosukhin}]{NIPS2017_3f5ee243}
Ashish Vaswani, Noam Shazeer, Niki Parmar, Jakob Uszkoreit, Llion Jones,
  Aidan~N Gomez, \L~ukasz Kaiser, and Illia Polosukhin. 2017.
\newblock \href
  {https://proceedings.neurips.cc/paper/2017/file/3f5ee243547dee91fbd053c1c4a845aa-Paper.pdf}
  {Attention is all you need}.
\newblock In \emph{Advances in Neural Information Processing Systems},
  volume~30. Curran Associates, Inc.

\bibitem[{Wang and Lu(2020)}]{wang2020two}
Jue Wang and Wei Lu. 2020.
\newblock \href {https://doi.org/10.18653/v1/2020.emnlp-main.133} {Two are
  better than one: Joint entity and relation extraction with table-sequence
  encoders}.
\newblock In \emph{Proceedings of the 2020 Conference on Empirical Methods in
  Natural Language Processing ({EMNLP} 2020)}, pages 1706--1721. Association
  for Computational Linguistics.

\bibitem[{Zhang et~al.(2020)Zhang, Jiang, Miura, Manning, and
  Langlotz}]{zhang2020contrastive}
Yuhao Zhang, Hang Jiang, Yasuhide Miura, Christopher~D Manning, and Curtis~P
  Langlotz. 2020.
\newblock \href {https://arxiv.org/abs/2010.00747} {Contrastive learning of
  medical visual representations from paired images and text}.
\newblock \emph{arXiv preprint arXiv:2010.00747}.

\bibitem[{Zhao et~al.(2020)Zhao, Hu, Cai, and Liu}]{zhaomodeling}
Shan Zhao, Minghao Hu, Zhiping Cai, and Fang Liu. 2020.
\newblock \href {https://doi.org/https://doi.org/10.24963/ijcai.2020/558}
  {Modeling dense cross-modal interactions for joint entity-relation
  extraction}.
\newblock pages 4032--4038.

\bibitem[{Zhao and Grishman(2005)}]{zhao2005extracting}
Shubin Zhao and Ralph Grishman. 2005.
\newblock \href {https://doi.org/https://doi.org/10.3115/1219840.1219892}
  {Extracting relations with integrated information using kernel methods}.
\newblock In \emph{Proceedings of the 43rd Annual Meeting of the Association
  for Computational Linguistics ({ACL}’05)}, pages 419--426.

\end{thebibliography}
\bibliographystyle{acl_natbib}

\end{document}